\documentclass[11pt, a4paper]{article}

\usepackage{graphicx}
\usepackage{amsfonts}
\usepackage{amssymb}
\usepackage{algorithm}
\usepackage{algorithmic}
\usepackage{subfig}

\newcommand{\dist}{\mathrm{dist}} 

\newtheorem{theorem}{Theorem}
\newtheorem{de}{Definition}
\newtheorem{con}{Conclusion}

\newenvironment{keywords}{
       \list{}{\advance\topsep by0.35cm\relax\small
       \leftmargin=1cm
       \labelwidth=0.35cm
       \listparindent=0.35cm
       \itemindent\listparindent
       \rightmargin\leftmargin}\item[\hskip\labelsep
                                     \bfseries Keywords:]}
     {\endlist}

\title{\textbf{A Survey on Open Problems\\for Mobile Robots}}
\author{\\Alberto Bandettini, Fabio Luporini, Giovanni Viglietta\\\\
			 {\small University of Pisa}}
\date{\today}

\begin{document}
\maketitle


\begin{abstract}
\textit{Gathering} mobile robots is a widely studied problem in robotic research. This survey first introduces the related work, summarizing models and results. Then, the focus shifts on the open problem of gathering \emph{fat} robots. In this context, \textquotedblleft fat\textquotedblright\ means that the robot is not represented by a point in a bidimensional space, but it has an extent. Moreover, it can be \emph{opaque} in the sense that other robots cannot \textquotedblleft see through\textquotedblright\ it. All these issues lead to a redefinition of the original problem and an extension of the CORDA model. For at most $4$ robots an algorithm is provided in the literature, but is gathering always possible for $n>4$ fat robots?

Another open problem is considered: \emph{Boundary Patrolling} by mobile robots. A set of mobile robots with constraints only on speed and visibility is working in a polygonal environment having boundary and possibly obstacles. The robots have to perform a perpetual movement (possibly within the environment) so that the maximum timespan in which a point of the boundary is not being watched by any robot is minimized.
\end{abstract}

\section{Introduction}\label{sec1}
The current trend in robotic research is to employ many simple robots that are capable, together, to perform rather complex tasks~\cite{8}. There are several advantages in keeping the power of the robots as low as possible: for instance, the cost or the system expandability. Hence, in this context, a robot is a weak entity in terms of computational capabilities. Different models have been introduced in the literature to represent mobile robots (such as the CORDA model or the ATOM model). Moreover, in some studies, robots may have additional features, like \emph{multiplicity detection}. This scenario shifts the research interest toward studying the different tasks that robots can perform from a particular standpoint: the algorithmic limitations with respect to the capabilities of the robots.

\emph{Pattern Formation} is a basic coordination problem for mobile robots, which consists in moving robots in order to form an arbitrary given pattern~\cite{8,39}. A special case is the \emph{Gathering} problem, also called in the literature \emph{rendez-vous}, \emph{F-POINT} or \emph{formation point}~\cite{1,2,3,4,8,9,10,11,12,13,14,22,35,39}. Informally, the aim is to gather all the robots in a point not chosen in advance. In spite of its apparent simplicity, several factors render this problem difficult to solve. In~\cite{8} a description of the current investigations shows how algorithmic solutions are strongly influenced by robot settings. In order to clarify the ideas, all these aspects will be treated in Section~\ref{sec2}, and then we will introduce the \emph{Gathering} problem with \emph{fat} robots, which is the focus of the first part of this survey. We will describe several potential approaches to tackle the problem, both new and based on the literature on the traditional Gathering problem.

In Section~\ref{sec3} a second problem is discussed: a team of mobile robots, having a limited visibility range, has to patrol a polygonal environment so that the maximum timespan in which a point of the boundary is not being seen by any robot is minimized. The \emph{Boundary Patrolling} problem is not new in the literature~\cite{40, 41, 42, 43, 44, 45, 46, 47, 48, 49, 50, 51, 54}, but it has never been treated with a proper algorithmic approach. Actually, another issue is the (almost total) lack of theoretical results for any given robot model; that is, the Boundary Patrolling problem has been tackled mainly from an empirical point of view. After summarizing the state of the art, we will give some insights on how the research on this problem could proceed.

\section{Gathering fat mobile robots}\label{sec2}

In this Section we will describe the robot model and the main results related to the problem of Gathering mobile robots. Then we will introduce the version of the problem concerning \emph{fat} robots and the recent investigations on it. Finally, we will present the open problem of Gathering fat mobile robots in an extended CORDA model: is gathering always possible for configurations of $n>4$ fat robots?

\subsection{The original model}
\label{model}
According to~\cite{1,2,3,8,9,10,11,12,13,14,22,35,39} we can summarize the model as follows. The robots are \emph{anonymous} entities viewed as \emph{points} in $\mathbb{R}^2$. \emph{Anonymity} means that all the robots are executing the same \emph{deterministic} algorithm and they have no way to \textquotedblleft recognize\textquotedblright\ the others, but are only able to detect their positions according to their \emph{local} coordinate system. Indeed, robots can have \emph{total}, \emph{partial} or \emph{no agreement} on the coordinate system: this means that the local view of a robot can differ from another one on unit length, origin, direction and orientation of the axes. Obviously, the robots can freely move in the plane and they are \emph{autonomous}, in the sense that they have to perform tasks in a totally \emph{distributed} manner (hence there is no central coordinator). Robots cannot communicate and their life cycle consists in executing an infinite loop of four phases: \emph{waiting}, \emph{sensing}, \emph{computing} and \emph{moving}. Initially all the robots are in the waiting state, but a robot cannot stay indefinitely idle. Hence, at a given time, it starts observing the positions of the other robots using its own sensors (sensing) and computes a destination point according to this information, the deterministic algorithm and the local coordinate system (computing). Optionally, some other information from the past can be used, but in this survey we are mostly focused on robots with no memory about previous observations and calculations (\emph{oblivious} robots). It may happen that a robot does not have \emph{unlimited visibility}. In this setting, a robot can detect another one only if it lies within a fixed finite distance $r$. The last phase consists in moving toward the computed destination, perhaps stopping before actually reaching it.

Further, robots can be \emph{asynchronous}, \emph{semi-synchronous} or \emph{totally synchronized}. 

In the first case (also called \emph{CORDA} or \emph{general setting}) robots may start each phase at any moment. Only two assumptions are made.
\begin{enumerate}
\item \textit{The execution time of a phase is not infinite}. The purpose is to ensure that Gathering does not become impossible because some robots take an infinite amount of time to complete a phase. Anyway, the duration of each phase is unpredictable. This also implies that a robot may see another one while it is moving, and the decisions taken accordingly may not be based on recent information, because the time between looking and moving phases can be arbitrarily long.
\item \textit{The distance traveled in a moving phase is not infinitesimal}. In other words, there is a fixed distance $d$ such that, if the computed destination of a robot is at distance greater that $d$, then the robot moves of at least $d$ in the next moving phase. Note that, without this assumption, it would be impossible for any algorithm to always terminate in finite time. Other than this, the distance actually traveled is unpredictable, because a robot may stop at any time before reaching its target.
\end{enumerate}

In the semi-synchronous case, at each turn there are \emph{active} and \emph{inactive} robots. All the active robots start a specific turn in a specific moment, while the others do nothing. In this setting there is total agreement on timing and, consequently, it is not possible for a robot to see another one while it is moving. For this reason, the movement of a robot can be considered \emph{instantaneous}, and this setting is also called \emph{ATOM}, since robots perform each turn in an \emph{atomic} fashion~\cite{35}. In contrast with the CORDA model, destinations are always calculated on recent observations. At different turns, different robots may active, but no robot may stay forever inactive.

The third case (totally synchronized) is a particular sub-case of ATOM in which all the robots are always active, hence there is total synchrony among them. 

The ability to detect the presence of more than one robot in the same point of the plane is called \emph{multiplicity detection}. As shown in the next Subsection, this feature gives rise to several variations on the problem.

Taking into account the above remarks, we can give a definition of the Gathering problem, as stated in~\cite{11}.

\begin{de}
\label{gather}
Given a set of robots arbitrarily placed in the plane, no two of them lying in the same location, Gathering consists in moving all the robots to a point in the plane not chosen in advance, in a finite amount of time.
\end{de}

It is worth stressing that Gathering must be solved in finite time, otherwise we are considering a different problem, called \emph{convergence point} (or \emph{C-POINT}), in which robots approach the gathering point but may never reach it. Important considerations are made for this problem too, and we will further explain them in the following, because they will be useful in our subsequent treatment of \emph{fat} robots.

\subsection{Main results}
\label{result}

In spite of its apparent simplicity, several factors make Gathering difficult to solve, as suggested by the following Conclusion, stated in~\cite{35}.

\begin{con}
In both the asynchronous and semi-synchronous settings, there exists no deterministic oblivious algorithm that solves Gathering for a set of $n\geqslant 2$ robots.
\end{con}

On the other hand, if we consider the C-POINT problem, there exists an algorithm that makes the robots converge to a point. As described in~\cite{12,13}, it is sufficient to move each robot toward the \emph{center of gravity} $c$, which is defined as
\[
c=\frac{\sum_{i=1}^n p_i}{n}\ ,
\]
where the $p_i$'s are the locations of the robots. The disadvantage of this solution is that the center of gravity is not \emph{invariant} and, as a consequence, this algorithm only achieves convergence. We recall from the definition of the problem that the gathering point is not chosen in advance. Hence this point is calculated by all robots taking into account some properties like the center of gravity above.  The \emph{Fermat-Torricelli} (or \emph{Weber}) point has important properties too and, further, it stays the same even after some robots have moved toward it. Unfortunately it is not computable by arithmetic operations and $k$-th roots if the number of robots is greater than a few units~\cite{10}. So this approach cannot be a general solution for the Gathering problem but, as it will be apparent when dealing with \emph{fat} robots, it can be very useful in a specific case.

Anyway, the above impossibility result can be prevented if the nature of the robots changes. If, for instance, they are able to detect multiplicity, then Gathering becomes solvable. As explained in~\cite{10}, the idea is to gather two robots in a point and, subsequently, move all the others toward this point. The important requirement is that the point with more that one robot remains unique during the entire execution of the algorithm.

\begin{con}
In both the asynchronous and semi-synchronous setting, there exists a deterministic oblivious algorithm that solves Gathering for a set of $n\geqslant 3$ robots, if multiplicity detection is available.
\end{con}

In no case it is possible to gather two robots. In~\cite{9}, a proof of this rather disappointing result is given, in a model where the robots have unlimited memory of past results and observations (\emph{non-oblivious} robots).

So far, we were assuming \emph{unlimited} visibility of the robots. A further improvement is to consider \emph{limited} visibility, which was studied mainly in~\cite{2,22}. In order to summarize the known results, it is convenient to define the \emph{visibility graph}, which is defined as the graph $\mathcal{G}$ on the robot set, where there is an edge between nodes $i$ and $j$ if and only if the robot $i$ belongs to the \emph{visibility disk} $D(j)$ of $j$, and vice-versa. Formally:

\begin{de}
\label{graph}
$\mathcal{G}(\mathcal{V}, E)\ s.t.\ \mathcal{V} = \lbrace p_1, \cdot\cdot\cdot, p_n\rbrace\wedge E = \lbrace (i, j) \mid p_i \in D(j) \wedge p_j \in D(i),\ i \neq j \rbrace$
\end{de}

It is important to notice that the radii $r_i$ and $r_j$ of $D_i$ and $D_j$ may differ, but both are finite. Clearly, Gathering can be eventually solved only if $\mathcal{G}$ is connected. Assuming the connectedness of $\mathcal{G}$,~\cite{2} shows a solution only for the C-POINT problem; whereas~\cite{22} solves Gathering only if the robots have total agreement on the coordinate system.
 
Further related work about Gathering is found in~\cite{14} and in~\cite{1}.  In the first paper the authors propose an algorithm for converging under the more realistic assumptions that robot sensors, movement and internal calculations may have slight \emph{inaccuracies}; while the latter investigates the behavior of the robots in the presence of \emph{failures}.

\subsection{Fat robots}

The model described in Section~\ref{model} is a good starting point for dealing with the problem, but it has several strong limitations. First of all, real robots are not points, but they have a physical extent. Moreover, incidental collisions between robots are neglected in this model, and generally avoided in the published algorithms. Finally, the presence of a robot may obstruct the visibility of other robots (robots may be \emph{opaque}, as opposed to \emph{transparent}). All the assumptions in the theoretical model have the possible consequence that real results (i.e., based on simulations with real robots) may significantly differ from the expectations. For example, an algorithm that works in CORDA may fail to work with opaque robots. Indeed, a robot cannot sense the position of robots hidden behind other robots, so it may not be able to always compute the proper destination point. It is easy to recognize that the model described in Section~\ref{model} must be extended, in order to include all these features.

For these reasons, research is oriented toward considering \emph{fat} robots, which are not points in the plane, but have an extension. A first definition, based on the work in~\cite{7}, may be the following.

\begin{de}
A fat robot is a robot with a physical extent representable as a unit disk embedded in $\mathbb{R}^2$.
\end{de}

One advantage over the previous scenario is the partial agreement on the coordinate system, that come for free in this model, since fat robots have all the same radius, which can be taken as the common \emph{unit length}.

As anticipated, considering the \textquotedblleft fatness\textquotedblright\ of the robots, two aspects become relevant, as explained in~\cite{7}.

\begin{itemize}
\item \textit{Presence of opaque robots}. Robot $r_i$ can see robot $r_j$ if there exist points $p$ and $q$ on the disks representing $R_i$ and $R_j$ respectively, such that the segment $pq$ does not contain any point of any other robot. A robot has \emph{full visibility} if it can see all the others.
\item \textit{Collisions between robots}. If a robot touches another one, then both robots stop and their moving phase ends (the model in~\cite{7} works in the asynchronous setting).
\end{itemize}

Since robots are fat and they stop moving in case of collision, it is not possible to gather them in a single point as before. For this reason, the very nature of the Gathering problem is different, compared to the definition given in Section~\ref{model}.

\begin{de}
\label{gatherfat}
Gathering fat robots consists in forming a configuration in which the union of all the disks representing robots is a connected set.
\end{de}

It is understood that the algorithm should also \emph{terminate}, in the sense that each robot has to become eventually aware that all the robots have gathered, and this task is not always straightforward. Indeed, starting from a configuration in which all the $n>2$ robots are \emph{collinear}, a Gathering configuration could be achieved by moving all the robots along this line, but then no robot would ever be aware that all is finished~\cite{7}, because its neighbors obstruct its view. So the problem reformulated in this way is more complicated than the one stated in Definition~\ref{gather} and, for solving it, we will assume that every robot knows the total number $n$ of the robots.

\begin{figure}[h]
	\centerline{
		\mbox{\includegraphics[width=0.8\linewidth , height=0.8\textheight , keepaspectratio]{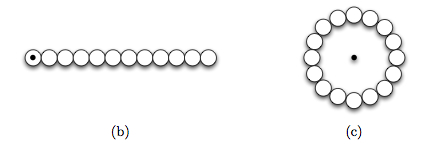}}
	}
	\caption{Some possible configurations of weak Gathering.}
	\label{weakgathering}
\end{figure}

Notice that, to solve Gathering as stated in Definition~\ref{gatherfat} (but disregarding termination), it is sufficient to form some highly sparse configurations, such as those in Figure~\ref{weakgathering}. Both configurations in the picture have diameter roughly proportional to the number of robots. In the following we will call this \emph{weak Gathering}, but we are not quite interested in it since, due to fatness, robots in final configurations can be very far apart, as also remarked in~\cite{6}. Thus, first of all, we want to improve the previous Definition.

Let the \emph{contact graph} of a configuration of fat robots be the planar graph on the set of their centers, with a straight edge connecting two centers if and only if their distance is exactly 2 (i.e., if the corresponding robots touch each other). Notice that weak Gathering configurations are precisely those whose contact graph is connected. We can now define \emph{strong Gathering} as follows:

\begin{de}
\label{stronggather}
Strongly Gathering $n$ fat robots consists in forming a configuration whose contact graph is connected and contained in a disk of radius $O(\sqrt n)$.
\end{de}

Of course, the distinction between strong and weak Gathering matters only when $n$ grows unboundedly. In this case, configurations such as those in Figure~\ref{weakgathering} do not achieve strong Gathering for large-enough $n$.

We could further improve the last definition by allowing only \emph{tightly packed} configurations. For instance, we could want the contact graph to be a 2-connected planar graph whose faces are equilateral triangles of side length 2 (except the outer face). In other words, we may want every bounded face of the contact graph to have exactly 3 vertices, and the union of the bounded faces to be a 2-manifold with boundary. Robots gathered in such a way form a \emph{rigid} configuration with no \emph{joints}. However, we are satisfied with Definition~\ref{stronggather}, as it already incorporates our intuitive notion of gathering.

Gathering one or two fat robots is trivial, and in~\cite{7} there are algorithms for three and four robots, too.

Let us begin with three robots. The solution proposed in~\cite{7} analyzes the two possible classes of configurations in which the three robots may be found: collinear and non-collinear, and takes action accordingly. The collinear case is simple, and the reader can refer to~\cite{7} for the details. In case the robots form a non-degenerate triangle, recall that they can be gathered in their Fermat-Torricelli point (see~\cite{10}). In~\cite{7} this strategy is adopted, but the computed destination point is actually $\frac{2\sqrt{3}}{3}$ away from the Fermat-Torricelli point, in such a way that the robots touch one another upon reaching it, without properly colliding.

From Section~\ref{result} we know that the Fermat-Torricelli point is invariant under movements toward it of the robots, but it is not computable in our model, unless the number of robots is very small. By definition, the Fermat-Torricelli point is the point that minimizes the sum of distances from the robots. For triangles, there exists an alternative characterization of the Fermat-Torricelli point (see~\cite{23}).

\begin{theorem}
If all the angles of the triangle $ABC$ are less than $120^{\circ}$, then the Fermat-Torricelli point $F$ satisfies $A\widehat{F}B=A\widehat{F}C=B\widehat{F}C=120^{\circ}$. Otherwise, it is the vertex corresponding to the greatest angle.
\end{theorem}

So the robots can compute this point easily. First of all, if some angle is greater than $120^{\circ}$, the Fermat-Torricelli point coincides with the corresponding vertex. Otherwise, the point can be computed as follows.

\begin{figure}[h]
	\centerline{
		\mbox{\includegraphics[width=0.5\linewidth , height=0.5\textheight , keepaspectratio]{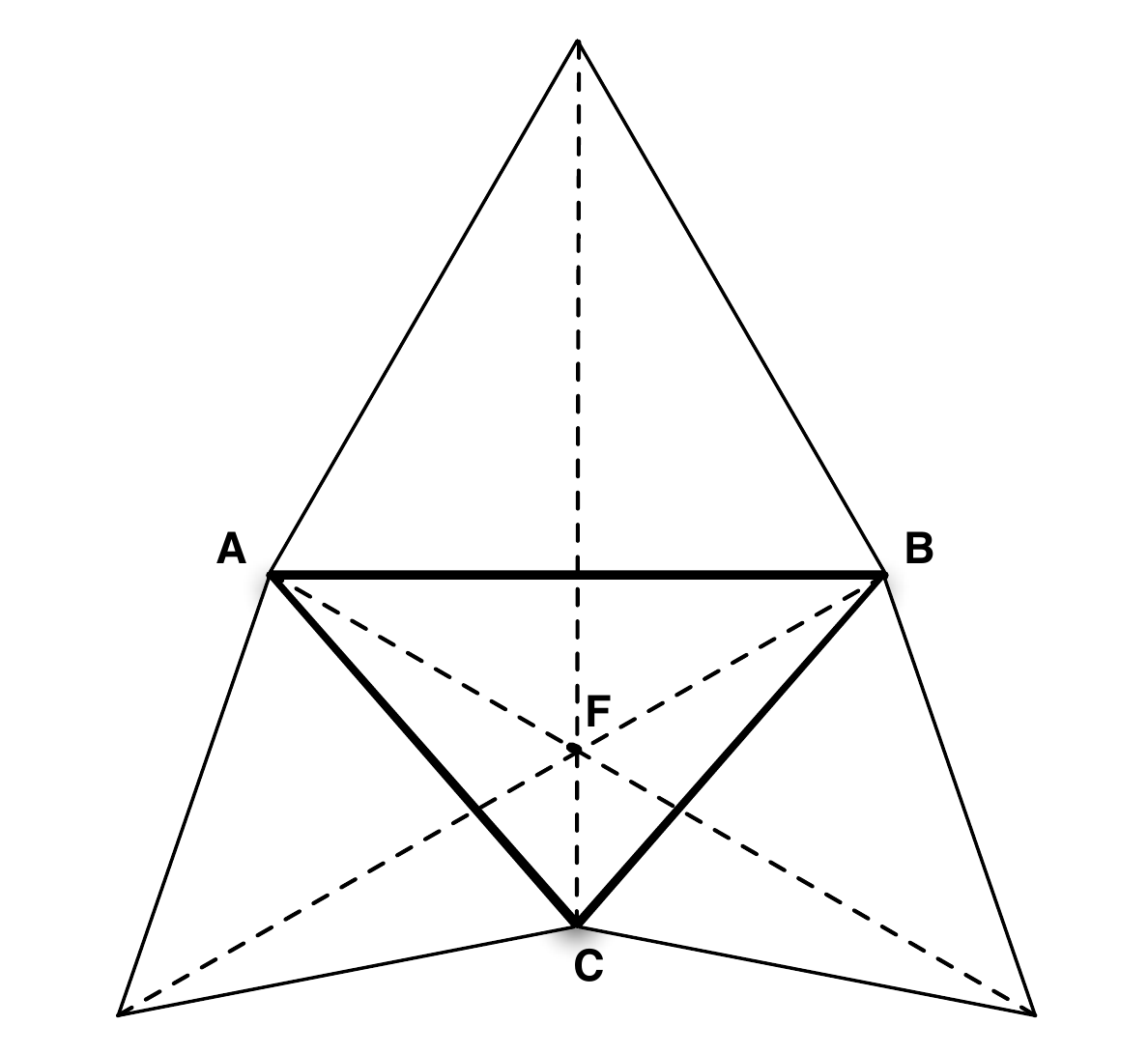}}
	}
	\caption{Construction of the Fermat-Torricelli point $F$ in a triangle $ABC$ whose angles are less than $120^{\circ}$.}
	\label{fermat}
\end{figure}

\begin{enumerate}
\item Construct three equilateral triangles on the three sides of the given triangle $ABC$, externally (refer to Figure~\ref{fermat}).
\item For each new vertex of the equilateral triangles, draw a line from it to the opposite triangle's vertex. 
\item These three lines intersect at the Fermat-Torricelli point, as Figure~\ref{fermat} shows.
\end{enumerate}

\begin{figure}[h]
	\centerline{
		\mbox{\includegraphics[width=0.5\linewidth , height=0.5\textheight , keepaspectratio]{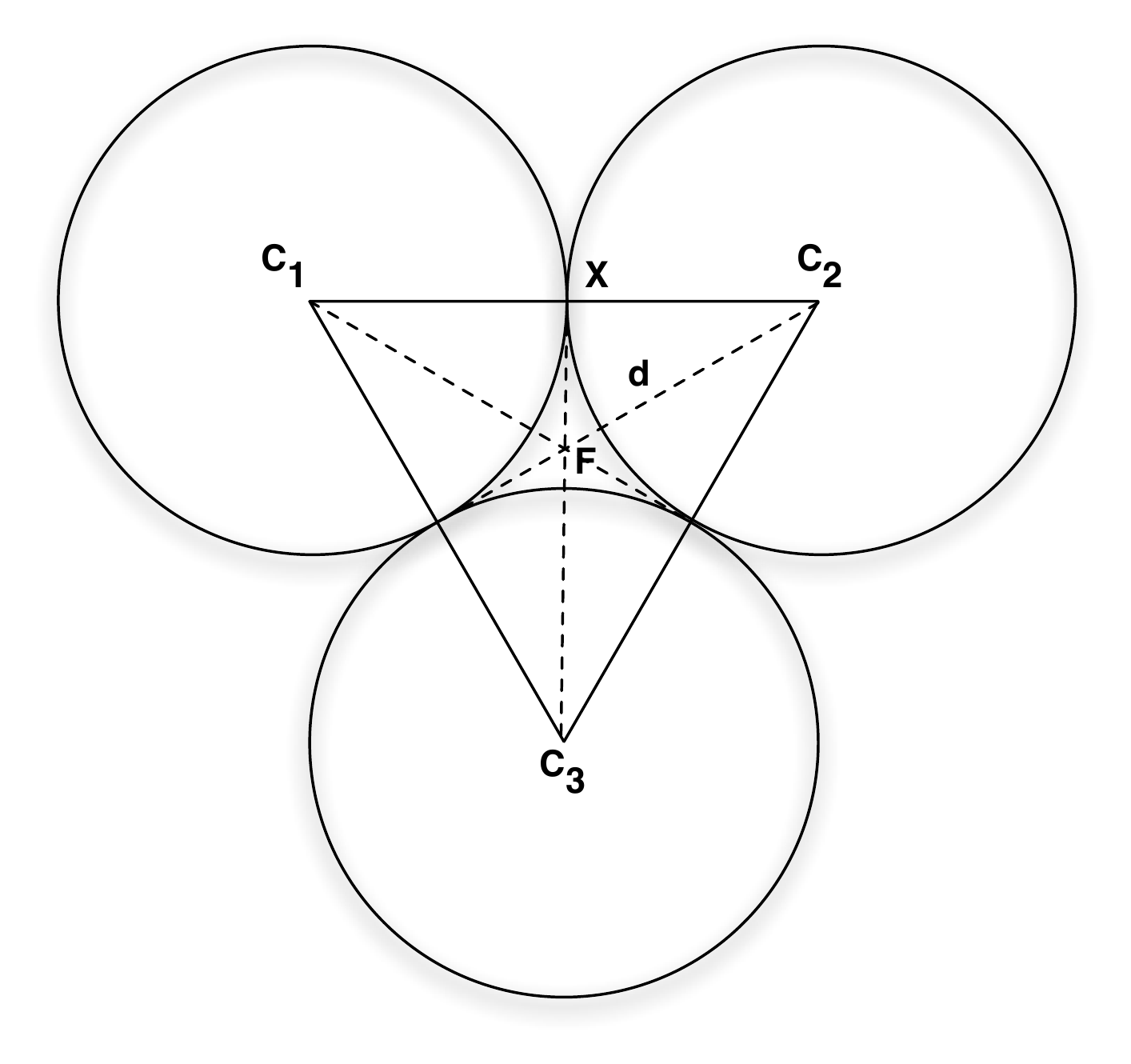}}
	}
	\caption{Gathering configuration for three fat robots.}
	\label{equilatero}
\end{figure}

Taking into account that the robots are fat, the destination point to be computed at each phase is not the Fermat-Torricelli point $F$, but it is a point at distance $\frac{2\sqrt{3}}{3}$ from $F$. The choice of this distance is justified by Figure~\ref{equilatero}, in which the three robots are in Gathering configuration. 

Triangle $C_{1}C_{2}C_{3}$ is equilateral with side length $l=2$. By elementary geometry, the computed distance is
\[
d = \frac{l}{\sqrt 3}= \frac{2\sqrt{3}}{3}.
\]

The other algorithm in~\cite{7} regards the gathering of four robots. The approach is \emph{brute-force}-like, in the sense that all the possible robot configurations are considered exhaustively, taking into account the actual views of the robots. Then, a different gathering strategy corresponds to each possible case. Of course, the same approach cannot be used when the number of robots grows, since the number of configurations quickly blows up.

In summary, this approach does not seem suitable to gather more than four robots, but the paper also suggests that achieving full visibility may be a way to solve the problem.

Indeed, the authors of~\cite{5} assume full visibility by considering transparent robots. The result is an algorithm for gathering $n>4$ transparent fat robots if and only if it is possible to \emph{break the symmetry} among them, in order to elect a \emph{leader}. In this algorithm, only one robot at a time can move toward the computed destination. Gathering is accomplished by forming a configuration with a robot in the centre of the \emph{Smallest Enclosing Circle}, and all the other robots arranged in a circular layered structure around it. This structure also satisfies our definition of strong Gathering.

We observe that the problem of electing the robot that has to move at each turn is the same \emph{Leader Election} problem studied in \cite{24, 25}.

\begin{de}
Leader Election consists in moving the system from an initial configuration where all entities are in the same state into a final configuration where all entities are in the same state, except one (the leader) that is in a different state.
\end{de}

\cite{25} shows that Leader Election and Pattern Formation are practically the same problem in CORDA with partial axis agreement and with $n\geqslant 4$. We recall from the Section~\ref{sec1} that Gathering is a particular Pattern Formation problem.

On the other hand, from the theory of design and analysis of distributed algorithms~\cite{24}, we know that the Leader Election problem is unsolvable if the entities do not have unique identifiers (although this holds for distributed systems that are not modeled by mobile robots).

So, how is it possible to elect a leader in a set of anonymous robots? According to the definition of anonymity in Section~\ref{model}, robots do not have identifiers and follow the same deterministic algorithm, but this does not mean that there is no way to sort them or, equivalently, to assign them a virtual ID. Indeed, it is possible to sort them according to their distance from the center of their Smallest Enclosing Circle $C$. Usually, the robot nearest to $C$ is selected. But the problem still arises in case more than one robot has minimum distance from $C$. In this case, an algorithm is provided for trying to break the symmetry but, if it is not possible, Leader Election is not performed, and the algorithm in~\cite{5} fails to achieve Gathering. However, operating in such a way, collisions are avoided because only the robot nearest to $C$ moves toward it, and no obstacles can ever appear in its path.

The continuation of this work, as the authors say in their conclusions, is to ensure that no symmetric configuration is formed during the execution of the algorithm and, in addition, to drop the assumption of transparent robots.

Another related study is presented in~\cite{6}. In this paper an algorithm is proposed for synchronous fat robots solving a problem that is very close to strong Gathering, as we know it from Definition~\ref{stronggather}, in that the aim is to gather fat robots as densely as possible in a circular area. An important difference is that the gathering point is fixed in advance, and is given to the robots as input, in contrast also with Definition~\ref{gather}. The paper discusses two variants: one in continuous space and time, and one in discrete space (essentially $\mathbb{Z}^2$) and time. The solutions for both variants are based on two operations: \emph{PULL}, for approaching the gathering point, and \emph{SPIN}, for avoiding collisions.

In the continuous setting, in case of intersection of the trajectories of two robots, SPIN would prevent them from colliding. As a drawback of this approach, the algorithm is not deterministic. Indeed, the robots rotate in a randomly chosen direction around the gathering point, and therefore are expected to move farther from their neighbors. Moreover, \emph{deadlock} configurations may occur, in which robots remain stuck. This happens mainly when the initial configuration has high symmetry.

On the other hand, in the discrete setting, deadlock can never occur and the algorithm, in contrast with the previous version, is deterministic. Unfortunately, it may still happen that several robots try to move to the same position of the grid at the same time. Since having multiple robots in the same position is forbidden, due to their fatness, it is necessary to break symmetries somehow. Unlike in~\cite{5}, in this paper some solutions are discussed. However, the proposed solutions extend the original model by using either unique identifiers, or direct communication between robots, or randomization.

\subsection{Further research}\label{sec24}

In this Section we have reviewed the state of the art on the Gathering problem for fat robots. In summary, we argued that~\cite{6} may not be very useful for solving Gathering of many fat robots, unless we somehow weaken the CORDA model by assigning further capabilities to the robots. We pointed out that computing the Fermat-Torricelli point is not a viable solution if the robots are too many, nor is exhaustively searching the configuration space, as in~\cite{7}. Recalling our original purposes to explore the interplay between robot capabilities and algorithmic solutions,  we believe that some ideas of~\cite{5,7} can be further exploited. Moreover, we observe that obtaining full visibility could be a fruitful starting point. In particular, this can be done in unlimited visibility if the robots are either transparent (like in~\cite{7}), or if they are arranged in such a way that any two robots are mutually visible. To our understanding, this arrangement can be achieved in basically two ways.

\begin{enumerate}
\item The robots are placed in configurations in which \textquotedblleft nobody hides anybody\textquotedblright . This could be difficult to obtain (especially if the robots do not know their number) and is not always guaranteed in an initial configuration.
\item The Gathering algorithm run by each robot uses only the positions of its closest neighbors. This way, the presence of hidden robots is irrelevant to the computation of the destination point. In principle, this applies also to limited visibility models.
\end{enumerate}

Another idea, still largely unexplored in the literature, would be making full use of \emph{collision effects}.

We know that computing the center of gravity of all $n$ robots is not always possible, due to visibility limitations. If each robot computes the center of gravity of all the robots that it can see, then different robots may compute different destination points, and Gathering may never be achieved. Even without visibility limitations, the center of gravity of the robot set may change while the robots move. As a result, different robots may still compute different destination points. Another issue is that this approach may fail to achieve strong Gathering for fat robots, but instead form weak Gathering configurations, like those shown in Figure~\ref{weakgathering}. Further, in the traditional model, collisions are never exploited and, when they happen, they can stop the robots or give rise to deadlocks. So collisions are not acceptable and the above papers suggest ways to avoid them. In our opinion, collisions should be revalued because they may be useful for obtaining Gathering in very natural ways.

Indeed, one could study what happens if the center of gravity approach is employed in conjunction with fat robots with \emph{weight}. Considering a more realistic scenario in which robots are \emph{masses}, we can accordingly consider \emph{forces} between robots. In this situation, allowing interactions between robots could be an advantage, and collisions could be sought and not avoided, in contrast with the traditional model. These forces could be modeled by vectors, and their interactions managed as usual, by computing vector sums.

In our opinion, there are two main reasons for seeking collisions.

\begin{enumerate}
\item Destination points calculated by robots with the center of gravity approach can differ as explained above. This implies that the robots may not be headed to the very same point, but their collisions may affect their original directions and perhaps make them converge.
\item A robot can be pushed by another robot: this could be very useful for achieving strong Gathering. Indeed, external robots push the internal ones and force them to fill the empty spaces, thus reducing the Gathering area.
\end{enumerate}

On the other hand, even when the robots are fat but \emph{weightless}, moving each robot to the center of gravity of the visible robots could be a terminating strategy. It may fail to terminate in finite time (but only converge) if the robots are dimensionless, but with fat robots this strategy should steadily reduce the Smallest Enclosing Circle, until at least weak Gathering is obtained (and possibly strong Gathering, depending on the collision model).

A radically different approach would be to endow fat robots with some \emph{communication capabilities}, in order to make up for their lack of vision (due to their opaqueness), and for the complexity of their maneuvers. For instance, each robot could carry a \emph{colored light} visible to the robots around it, which can be set to different colors during the robot's computing phase. In Section~\ref{patrol-our}, we will argue that assuming the ability to form a \emph{line pattern}, perhaps with a little explicit communication between robots, could be an ideal intermediate step for solving both weak Gathering and strong Gathering.
\section{The Boundary Patrolling problem by mobile robots}\label{sec3}
\subsection{The problem}
\label{patrol-problem}
A set of $n$ mobile robots is working in a polygonal environment (a polygon with possibly polygonal obstacles in $\mathbb{R}^2$) having boundary $P$. Each robot has its own speed, bounded by some constant, and has limited visibility. Informally, the task of the robots is to perform a \emph{perpetual movement} in order to protect or supervise $P$. The robots' motion may or may not be constrained to the polygon itself. The problem of \emph{Boundary Patrolling} is also known as perimeter or border patrolling, and is acquiring more and more interest in the scientific community, due to its applications in various practical scenarios, like intrusion detection. Boundary Patrolling derives from the problem of \emph{boundary coverage}, where robots have to cover a perimeter by minimizing the number of visits to each point (ideally, visiting it only once)~\cite{44}. On the other hand, patrolling is in some sense opposite, because it aims at maximizing the frequency at which each point of $P$ is monitored by at least one robot. We now want to formalize the Boundary Patrolling problem.

Let $R = \lbrace r_1, \cdots, r_n \rbrace$ be a set of mobile robots and let $P$ be a closed polygonal chain representing the boundary to be monitored. Each robot $r_i$, at time instant $t$, is located at some position $r_i(t) \in \mathbb{R}^2$ and has limited visibility. 
\begin{de} 
Let $\dist(p,q)$ be the function denoting the Euclidean distance between points $p$ and $q$, and let $d > 0$ be a constant. The \emph{Visibility Range} of a robot $r_i$ at time $t$ is given by
$V_i(t) = \lbrace p \in \mathbb{R}^2 \mid \dist(r_i(t),p) \leqslant d \rbrace$.
\end{de}
Now we want to define the concept of monitoring a point. We would like also to consider the presence of \emph{obstacles} in $\mathbb{R}^2$, namely solid objects that occlude the vision of a robot. For instance, the boundary itself could limit the visibility range. If we want to model the presence of obstacles, we can introduce a set $O = \lbrace o_1, \cdots, o_m\rbrace$, $o_j \in \mathbb{R}^2$, such that each point $o_j$ represents a solid object in the plane. This way, we can state whether a point is monitored or not.
\begin{de}
A point $p \in \mathbb{R}^2$ is said to be \emph{monitored} at time $t$ if $\exists r_i \in R,\ p \in V_i(t)\ \wedge\ O \cap r_i(t) p = \varnothing$.
\end{de}
Whenever we assume absence of obstacles, we will simply say that a point is monitored at time instant $t$ if it lies in the visibility range of at least one robot.

For convenience, we introduce also the following notation.
$$
M(t) = \lbrace p \in \mathbb{R}^2 \mid p\ \mbox{is\ monitored at time}\ t\rbrace.
$$
We informally said that mobile robots have to patrol a boundary. Patrolling is the action of cyclically moving along a specific route in order to monitor the points of $P$.
\begin{de}
The \emph{perpetual movement} of a robot $r_i$ is a $k_i$-tuple $\Gamma_i = (p_1, \cdots, p_{k_i} )$, with $p_j\in \mathbb{R}^2$, describing the path of $r_i$.
\end{de}
In general, since the robots move, we will have $M(t) \neq M(t+\varepsilon)$. In this context, it is useful to introduce the following set.
$$
U(t_1, t_2) = \bigcup_{t \in [t_1, t_2]} M(t).
$$
$U(t_1, t_2)$ represents the set of points that have been monitored by at least one robot between time $t_1$ and time $t_2$. 

We have all the tools to define the goal of the Boundary Patrolling problem, which is to find the minimum time period $T$ and a perpetual movement for each robot $r_i \in R$ such that
\begin{equation}\label{patrol-goal}
\forall t \geqslant 0,\ P\subseteq U(t, t+T).
\end{equation}

\subsection*{State of the art}
Presently, as far as we know, there are no papers that specifically \emph{solve} the problem stated above. However, it is not hard to find material concerning the related problem of \emph{patrolling inside closed areas} by mobile robots (or agents), e.g.,~\cite{40, 41, 42, 43, 44, 45, 46, 47, 48, 49, 50, 51, 54}. We recall that it is common to talk about \emph{agents}, rather than \emph{robots}, whenever the environment in which the entities act is modeled as a \emph{graph}, as opposed to $\mathbb{R}^2$. Hence, to avoid any confusion, we remark that we are looking for solutions to the Boundary (not area) Patrolling problem performed by mobile robots (not agents).

There are also a few particular studies, about which we are going to talk in Section~\ref{patrol-ref}, that are somehow related to the Boundary Patrolling problem by mobile robots (e.g.,~\cite{40,41,42,50}): they are all good starting points for our purposes, but, for different reasons, they cannot be considered real solutions to our problem. Indeed, we anticipate that:
\begin{itemize}
\item in general, these investigation are not undertaken with an algorithmic flavor (e.g.,~\cite{41,42});
\item as a consequence, there are no proofs establishing the optimality of the patrolling algorithm, as requested by~(\ref{patrol-goal}), neither for arbitrary patters nor for specific geometric patterns (e.g., line segments, regular polygons, etc.).
\end{itemize}
Further, we must acknowledge the lack of a unified model to describe the capabilities of the robots. With respect to many other contexts, in which robots are assumed to operate under the well-known (and properly formalized) CORDA model~\cite{52}, we observe that studies on patrolling problems make several very different assumptions on the abilities of robots and the features of the environment. Our first step in order to start studying the Boundary Patrolling problem will be to formalize a model aimed at defining both capabilities and limitations of robots.   

\subsection{Modeling mobile robots}
We roughly said that robots are mobile entities with computational capabilities that freely move in $\mathbb{R}^2$. A robot can have more or less \textquotedblleft power\textquotedblright, in the sense that it may or may not have certain capabilities: for instance, it may be able to directly communicate with other robots within its visibility range, or explicit communications may be forbidden. Obviously, depending on the assumptions on the model, different solutions to our problem may be determined. In this Section we explain how to model a robot and which the capabilities that determine its power are. We remark that it would be interesting to bound the \emph{optimal patrolling period} $T$ (see~(\ref{patrol-goal})), assuming only a \emph{minimal set} of robot capabilities. That is, we may be interested in understanding the relationship between the power of the robots and the optimal solutions to the Boundary Patrolling problem. As stated in Section~\ref{sec1}, the trend in robotic research is to design and use a large number of low cost and very simple \textquotedblleft general-purpose\textquotedblright\ robots (the so called \textquotedblleft weak robots\textquotedblright, having a few capabilities), rather than a few, usually expensive, application-specific robots~\cite{52}. 

\paragraph{The environment.}
We first describe the polygonal environment in which mobile robots act. We assume that robots have \emph{global knowledge} of the environment; otherwise, other algorithms performing a preliminary exploration have to be employed, but this is not our matter of interest. Having global knowledge of the environment basically means that:

\begin{itemize}
\item $P$ is \textquotedblleft known\textquotedblright\ to all robots; this does not necessarily mean that robots can determine their position in $\mathbb{R}^2$ with respect to $P$, but only that they know the shape of the boundary.
\item Robots are aware of the presence of obstacles within the environment; this also means that a robot is able to distinguish obstacles from other robots. However, the environment may also be free of obstacles.
\end{itemize}

If one of these basic informations is unknown to the robots, then we are dealing with \emph{partial knowledge} of the environment. 

Robots do not usually know in advance the value of $n$ (e.g.,~\cite{40,41}); this is quite obvious and in some sense necessary to obtain algorithms that are as flexible as possible. The environment can be \emph{static} or \emph{dynamic} (see~\cite{22}). A dynamic environment can change during time: newer obstacle may appear, $P$ can change forcing robots to determine another optimal patrolling path, and so on. However, the environment is typically assumed to be static and globally known to all robots.

Finally, notice that $P$ could be either a specific geometric shape (regular polygon, line segment, etc.) or arbitrary: this is quite important because, for a given robot model, specific solutions to the Boundary Patrolling problem could be developed just for some patterns, like has been done for the Pattern Formation problem~\cite{52}.

\paragraph{Behavior of the robots.}
In our model robots are \emph{anonymous} entities that cyclically move in $\mathbb{R}^2$ in order to satisfy~(\ref{patrol-goal}).
The goal is to define an optimal \emph{patrol set} $\sigma = (\Gamma_1, \Gamma_2, \cdots, \Gamma_n)$ such that~(\ref{patrol-goal}) holds. There are two generic ways to approach the problem.
\begin{itemize}
\item \emph{Centralized setting}. A centralized authority knows the number of robots, their respective speeds and has global knowledge of the environment. It is in charge of computing $\sigma$.
\item \emph{Distributed setting}. Robots can exploit certain capabilities to execute a distributed algorithm. In this case, since the robots are anonymous, they all follow the same algorithm. $r_i$'s execution implicitly determines its own $\Gamma_i$.
\end{itemize}
We now focus on modeling a robot in a distributed setting, which is the case of our interest. Our robot model may be very different from the one that we formalized in Section~\ref{model}. This is mainly because it is still an open problem to determine a trade-off between a set of minimal robot capabilities and substantial solutions to the Boundary Patrolling problem. Further, in the literature no reference model has been proposed for patrolling robots yet, like somewhat happened for the CORDA model~\cite{52}.

In our model, each robot performs a local computation described by a distributed, possibly deterministic, algorithm. Thus, robots are said to be \emph{autonomous}. The dichotomy between \emph{oblivious} and (\emph{un})\emph{bounded memory}, and the way robots agree on a coordinate system, as stated in Section~\ref{model}, are still fundamental aspects of the model. On the other hand, we now assume that robots can sense the environment only within their visibility range. Their movement, described in terms of a common \emph{unit length}, is characterized by \emph{bounded speed}. The speed itself, which could vary from robot to robot, should be properly modeled: for instance,~\cite{50} proposes an accurate movement model in which different actions, like turning and changing direction, may have different costs in time.

Robots can operate in \emph{cycles} of \emph{stages} (e.g.,~\cite{50,52}) leading to either a \emph{synchronous} or an \emph{asynchronous} model, depending on the timing constraints on each stage. Typically, each stage represents an action performed by a robot (look, move, compute, etc.). However, other approaches have been proposed in the literature (e.g.,~\cite{40, 41, 42, 47}): we will comment them in Section~\ref{patrol-ref}.

\emph{Communication} between robots is also a key aspect of the model, because it has a strong influence on the power of robots and on the feasibility of the model. Communication can be \emph{explicit} or \emph{implicit}. It is explicit when a robot is able to deliver a message to some other robot. This way of exchanging informations can be \emph{global} or \emph{local}: in the first case, a robot can directly communicate with any other robot in $\mathbb{R}^2$, while in the second one, which is the most common in the literature, communication is restricted to the visibility range of each robot (e.g.,~\cite{40}). Explicit global communication gives robots a lot of power, but this is very costly, and sometimes infeasible in a practical implementation. On the other hand, implicit communication means that a robot acquires information by sensing the position of other robots within its visibility range (like in CORDA~\cite{52}); such information, together with the knowledge of the environment, would be the only data on which the patrol strategy is based (e.g.,~\cite{41}).

\subsection{Solutions and limitations}
\label{patrol-ref}
In this Section we summarize the main results achieved on the Boundary Patrolling problem, with particular emphasis on the limitations of such studies.

\paragraph{From Boundary Coverage to Boundary Patrolling.}
In~\cite{45} a detailed analysis of the Boundary Coverage problem is proposed. The problem is properly formalized and the robot model is clear. It may be interesting to understand the Boundary Coverage problem because of its strong connections with the Boundary Patrolling problem. For instance, as shown in~\cite{43}, some ideas for area coverage may be readopted for area patrolling. The weakness of studies like~\cite{43,45,48} lies in the methodology used to perform the coverage/patrolling, which basically relies on a graph-based strategy for path planning. Indeed, as pointed out in a more recent paper~\cite{40}, these approaches make a lot of assumptions on the robot model. For instance, they require the same data structure (representing the graph) to be shared by all the robots; this is very unlikely in real scenarios, especially when robots have to patrol an arbitrary pattern: a robot would have to act as leader and then build a representation of the graph, which finally would be sent to all the other robots by explicit communication.

\paragraph{Multi-agent patrolling.}
In~\cite{49, 54} an interesting analysis can be found on a multi-agent architecture for \emph{area} patrolling; even if it is not exactly related to our problem, these papers become important for a detailed taxonomy of the robot models. Aspects like perception, communication and coordination are addressed as first-class citizens, and their implications on the feasibility of the models are pointed out. In particular,~\cite{49} is one of the first attempts to approach the patrolling problem with agents that do not use explicit communication. In~\cite{51}, the contents of~\cite{49} are resumed and analyzed under a more algorithmic perspective: for the first time, various classes of patrolling strategies are compared with the theoretically \textquotedblleft optimal\textquotedblright\ one (executed by extremely powerful agents) by means of a standardized complexity analysis. Again, the practical problem lies in the assumptions on the agent model, which are, in the best case, a little less demanding than in~\cite{45}. 

\paragraph{Patrolling under frequency constraints.}
In~\cite{44} the focus is on designing a patrolling algorithm that guarantees \emph{maximal uniform frequency}, that is, each point in the \emph{area} has to be covered at the same optimal frequency. This is obviously strictly related to the goal of the Boundary Patrolling problem. The model is even complicated by the fact that robots have \emph{speed limitations} depending on both their traveling direction and the environment. In spite of this constraint, a lot of simplifications come from other assumptions: the environment is modeled as a graph (exactly like in~\cite{43,45,48}) whose edges have a \emph{weight} representing a speed reduction. A minimum Hamiltonian cycle is then computed on this graph, and all the robots uniformly spread along it. Finally, all robots travel either clockwise or counterclockwise to guarantee maximal uniform frequency (this behavior is quite intuitive but is also formally studied in~\cite{44}). We could attempt to readopt the same approach for the Boundary Patrolling problem, but unfortunately the weaknesses of this model are resounding. These robots are very powerful, and they act like agents (since they work in a graph) as opposed to proper robots. A leader has to be elected somehow, to build the graph and to compute the minimum Hamiltonian cycle. In order to do this, the leader has to be equipped with special tools to make a mapping of the environment. Then, the Hamiltonian cycle has to be explicitly communicated to the other robots. Furthermore, it is implicitly assumed that robots can locate themselves within the environment in order to take position uniformly around the boundary during the initialization phase; in this context, the initial assignment of locations to robots is computed offline by a central unit.

Now we focus on more recent papers (they appeared between 2009 and 2010) that have as main objective the study of boundary patrolling by multi-robot systems. Unfortunately, the lack of a proper algorithmic approach fades the relevance of these studies, at least for what concerns our purposes.

\paragraph{The AI perspective.}
In~\cite{41, 42} the goal is to investigate the Boundary Patrolling problem by mobile robots with an approach that is closer to Artificial Intelligence and robotics, and is based on the so called \emph{Behavioral Control} of robots. A finite set of \emph{elementary actions} that a robot can undertake (e.g., \textquotedblleft go to frontier\textquotedblright, \textquotedblleft patrol boundary\textquotedblright, and so on), is defined in a framework aimed at describing the behavior of robots. Then, a \emph{supervisor} is tasked to implement the \textquotedblleft action selection mechanism\textquotedblright. We could say that the supervisor, based on the sensorial capabilities of a robot, defines its way of reacting to the situations encountered in the environment, thus decomposing complex tasks into smaller and simpler sub-tasks. In a sense, Behavioral Control of robots is just an AI-like way of expressing and solving a problem involving multiple robots, which however suffers the lack of a theoretical analysis. In~\cite{41, 42} a team of robots has to patrol an arbitrary pattern with no obstacles, where robots have global knowledge of the static environment. Furthermore, as usual, robots can sense the boundary as soon as it appears in their visibility range. Robots are asynchronous, autonomous, oblivious, they have the same speed and their communication is implicit. The main drawback of the robot model is the capability of each robot of locating its own position within the environment. The difference between~\cite{41} and~\cite{42} lies in the supervisor: while in~\cite{42} a Finite State Automaton is adopted, in~\cite{41}, instead, a Fuzzy Inference System~\cite{FUZZY} is used to implement the action selection mechanism. However, in both cases, the algorithm describing the behavior of a robot is straightforward: exploiting its capabilities, a robot has to stay as close as possible to the boundary, avoid team mates and, when necessary, change direction. No assumption on the initial configuration of the robots is made. Experiments have been performed in both~\cite{41} and~\cite{42} but, unfortunately, they are not compared to any other result in the literature. Obviously, these approaches cannot be considered solutions to the Boundary Patrolling problem.

\paragraph{Boundary Patrolling with explicit communications.}
In~\cite{40}, the model of~\cite{41, 42} changes in two key aspects: there is no \emph{a priori} way for a robot to know its location within the environment, but explicit local communication is allowed. Hence, even if an important ability is lost, it is compensated by the possibility of communicating. In~\cite{40}, robots have again the same speed and have to patrol an arbitrary pattern. Once a robot reaches the boundary at some point $p$, it keeps patrolling around $p$, first moving clockwise by some length $X$, and then counterclockwise by the same length. The value of $X$ is then updated according to some heuristics; if a robot $A$ meets another robot $B$ while patrolling, they first exchange their $X$ value and then proceed patrolling in the opposite direction. The value of $X$ is updated by averaging $X_A$ and $X_B$. The idea behind this algorithm is to let robots converge to a final configuration in which the length of the boundary that they patrol is equal for all the robots. Only experimental results, showing the convergence of the algorithm, are provided. The emphasis of the paper is on the computational simplicity of the heuristic employed.

\paragraph{Randomness in Boundary Patrolling.}
In~\cite{50}, a team of robots have to patrol a boundary in presence of an adversary attempting to break through. The adversary is modeled as a daemon, with global knowledge of the environment, and knowledge of the patrolling scheme of the robots. The key observation of the paper is that if the patrolling algorithm is deterministic, then the adversary can certainly penetrate the boundary as soon as some point is outside the visibility range of the robots. Hence, the idea is to develop a \emph{non-deterministic} algorithm to patrol the boundary; the goal here becomes minimizing the probability of border penetration. Another outstanding feature of~\cite{50} is the algorithmic approach, which is analyzed also from a theoretical point of view.

\subsection{Further research}
\label{patrol-our}
We suggest some directions for further research, in the algorithmic perspective adopted in~\cite{53}, using definitions and terminology introduced in Section~\ref{patrol-problem}. We start discussing the less interesting case of centralized settings, then we will focus on distributed settings.

\subsubsection*{Centralized settings}
A \emph{Central Authority} $CA$ knows $P$, executes an algorithm $\Lambda$ to compute $\sigma = (\Gamma_1, \cdots, \Gamma_n)$, and instructs each robot $r_i$ to move according to $\Gamma_i$. Optimal $\sigma$'s are those that minimize $T$ in~(\ref{patrol-goal}).

The base case is when $CA$ can place robots at \emph{fixed} positions in such a way that the whole boundary $P$ is monitored at a single time $t$. In this case we talk about Boundary Coverage, rather than Boundary Patrolling. In Computational Geometry, a similar problem, called the \emph{Art Gallery problem}, has been studied, where $P$'s sides act as opaque walls (i.e., obstacles to visibility).

On the other hand, concerning the general case in which the $n$ robots have to patrol the boundary along some routes, a related problem in Computational Geometry is the \emph{Watchman Route problem}: here, the objective is to compute the shortest route a single watchman should follow in order to guard an entire area, given only the knowledge of $P$~\cite{58}. Again, $P$'s boundary acts as an opaque wall. Many variants of the problem have been proposed: for instance, the \emph{Watchman Route with Limited Visibility}~\cite{55} or the \emph{Multiple Watchmen Route problem}~\cite{56,57}, which however are based on heuristics and do not guarantee shortest routes. Part of the results in the literature regarding the Watchman Route problem can be reused for our purposes. Indeed, one of the first steps should be to understand under which assumptions and to what extent the Watchman Route theory could be useful. For instance, given a boundary $P$, assume that thanks to a specific Watchman Route algorithm (with limited visibility) we can determine a single robot's shortest route $\Gamma$ to visit $P$; assume also that $\Gamma$ is a closed path and that all our mobile robots move synchronously with equal speed. Then it is natural to conjecture that, on some broad classes of polygons/boundaries (including, for instance, regular polygons), the Boundary Patrolling problem is solved by computing $\Gamma$ and then uniformly spreading all the robots along it, making them move in the same direction (say, clockwise). The challenge here is determining exact conditions under which this solution is indeed optimal.

\subsubsection*{Distributed settings}
As pointed out in Section~\ref{patrol-ref}, the two main drawbacks of the state of the art for distributed solutions to the Boundary Patrolling problem are:
\begin{itemize}
\item the lack of a proper algorithmic approach (but mainly empirical approaches);
\item very often, the adoption of unrealistic robot models (\textquotedblleft powerful robots\textquotedblright).
\end{itemize}
Our objective is to study the problem with the algorithmic flavor that characterizes works on various distributed coordination problems. For instance,~\cite{52} offers a survey on the computational results concerning what a set of autonomous mobile robots can or cannot achieve. In~\cite{52} mobile robots follow the CORDA model and the emphasis is on some basic geometric problems, known as \emph{Pattern Formation}, \emph{Gathering}, \emph{Following and Flocking}. As a matter of fact, we would like to understand if these geometric problems can be used as \textquotedblleft building blocks\textquotedblright\ for the Boundary Patrolling problem. Indeed, we know that, given a certain model, many classical problems in distributed computing (e.g., Leader Election in CORDA) can be reduced to geometric problems (e.g., Asymmetric Pattern Formation in CORDA). For instance, assume that our mobile robots have sufficient capabilities to form a certain pattern $X = (p_1, \cdots, p_n)$ and assume also that robots have global knowledge of the environment, so they have to patrol a boundary $P$ (we recall that robots know the shape of $P$, but in general they are not aware of their positions within $P$). What if we were able to \emph{map} $X$ onto $P$, in such a way that robots can (optimally) patrol $P$ by cyclically moving from $p_i$ to $p_{i+1}$?

\begin{figure}[h]
	\centering
  	\subfloat[Initial configuration]{\label{circle-unformed}\includegraphics[scale=1.1]{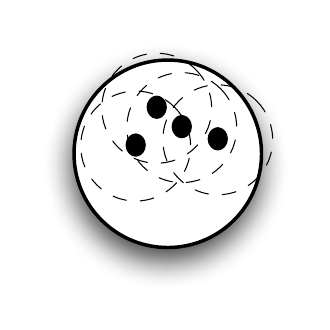}}  
	\hspace*{20pt}
  	\subfloat[Regular $n$-gon formed]{\label{circle-formed}\includegraphics[scale=1.1]{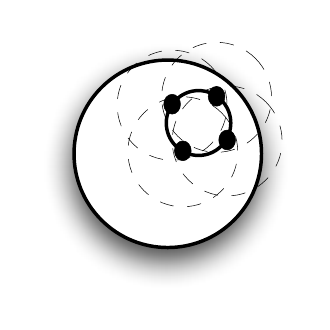}}  
  	\hspace*{20pt}
  	\subfloat[$n$-gon mapped onto $P$]{\label{circle-boundary}\includegraphics[scale=1.1]{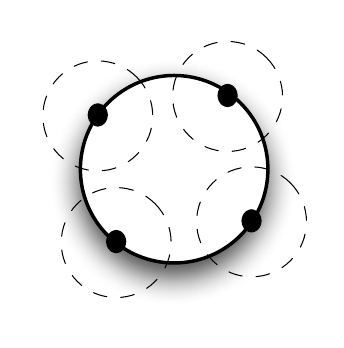}}  
	\caption{Regular $n$-gon formation in order to patrol a circular boundary.}
	\label{circle-patrol}
\end{figure}

Figure~\ref{circle-patrol} shows an example: $P$ is a circle, robots execute a distributed algorithm for regular $n$-gon Pattern Formation, then get onto $P$ and start moving in the same direction. This could be a way to solve the Boundary Patrolling problem in the special case in which $P$ is a circle, provided that a suitable robot model is employed. Obviously, pulling off this strategy is not simple: once the regular $n$-gon is formed, how can robots get uniformly around $P$? Once they are all on $P$, how do they coordinate their movements? Perhaps, in order for our strategy to work properly for reasonable robot models, we have to stipulate that $n$ is so large that the visibility graph of the robots (see Definition~\ref{graph}) remains connected throughout the execution of the algorithm, or a substantial part of it.

Notice that knowing the optimal patrolling route for specific boundaries could be useful even for other purposes. For instance, the optimal patrolling route for more complex environments (where the boundary is highly irregular, some rooms are narrower than the visibility range, there are several obstacles, etc.) could be approximated by composing algorithms for specific simple boundaries.

In CORDA, if the robots commonly agree on a coordinate system, then they can form any given pattern~\cite{53}. Forming certain patterns could be useful not only to solve the Boundary Patrolling problem like we suggested above, but also to obtain an agreement on \emph{roles} for subsequent, coordinated actions. For instance, we may devise a \emph{hybrid scheme}, in which some robots stand around a specific position of the environment, slightly moving in order to coordinate the motion of the other robots, who actually patrol the boundary.

Notice that all these \emph{multi-step} approaches, where each step corresponds to a different algorithm to run, in general require some kind of \emph{memory} on the robots, such as the colored lights we mentioned in Section~\ref{sec24}.

Additionally, a fundamental assumption of~\cite{53} cannot be made in our robot model: unlimited visibility. To the best of our knowledge, except for Gathering, all the available algorithms that are somehow related to Pattern Formation in CORDA assume unlimited visibility. If we hold to the idea of approaching the Boundary Patrolling problem by exploiting the Pattern Formation theory, then we could proceed in one of the following ways.
\begin{enumerate}
\item We assume that, in the initial configuration, the visibility graph is complete. In this case, the theory on Pattern Formation can be employed as it is, since we can pretend to be in an unlimited visibility setting. However, this is likely to be an unrealistic assumption, or at least a substantial constraint on the model.
\item We assume that the visibility graph is connected and we give robots some communication capabilities.
\item We study new distributed algorithms to form specific patterns in the limited visibility setting.
\end{enumerate}
In this context, an especially desirable pattern to form would be the \emph{line segment}. Once a line is formed, a proper Following and Flocking algorithm could be studied to let robots locate and reach the boundary while keeping their relative positions in the formation. A (possibly suboptimal) solution to the Boundary Patrolling problem would be to uniformly spread robots along $P$, so that the boundary is evenly divided into $n$ segments of equal length, each patrolled by a single robot from left to right and vice-versa. Alternatively, all the robots could walk along the boundary in the same direction (say, clockwise) at roughly the same speed.

This kind of strategy poses several issues. How can robots understand that a line pattern is formed, and that it is time to move on to the next phase? Some sort of communication is perhaps needed here. How can robots coordinate in order to end up at equal distances on the boundary? We do not know if a self-stabilizing algorithm can be found for this, or if we have to rely once again on communications. If we do not want to allow explicit communications, we could trade that ability for some additional robots, who could form a central \emph{coordination system}, by indirectly communicating with the boundary robots and synchronizing them.

As a side note, we observe that obtaining a line pattern with fat robots (refer to Section~\ref{sec2}) would not only achieve weak Gathering in a trivial way, but possibly also strong Gathering, via a similar Following and Flocking strategy.


\clearpage
\addcontentsline{toc}{section}{References}

\bibliographystyle{plain}	
\bibliography{biblio}		

\end{document}